\documentstyle[latex-acl,named]{article}

\title{Learning Transformation Rules to Find Grammatical Relations\thanks{\hspace{1em}This paper reports on work performed at the MITRE Corporation under
the support of the MITRE Sponsored Research Program. Helpful
assistance has been given by Yuval Krymolowski, Lynette Hirschman and an anonymous
reviewer. Copyright \copyright 1999 The MITRE Corporation. All rights reserved.}}
\author{Lisa Ferro \and Marc Vilain \and Alexander Yeh\\The MITRE Corporation\\ 202 Burlington Rd.\\ Bedford, MA 01730\\USA\\\{lferro,mbv,asy\}@mitre.org}

\begin{document}
\maketitle
\begin{abstract}
{\begin{picture}(0,0)
\put(0,215){Appears in Computational Natural Language Learning (CoNLL-99), pages 43-52.}
\put(10,200){A workshop at the 9th Conf.\ of the European Chapter of the
Assoc.\ for Computational Linguistics (EACL-99).}
\put(10,185){Bergen, Norway, June, 1999.}
\put(0,-485){cs.CL/9906015}
\end{picture}}
Grammatical relationships are an important level of natural
language processing. We present a trainable approach to find these
relationships through transformation sequences and error-driven
learning. Our approach finds grammatical relationships 
between core syntax groups and bypasses much of the parsing
phase. On our training and test set, our procedure achieves 63.6\%
recall and 77.3\% precision (f-score = 69.8).
\end{abstract}

\section{Introduction}
An important level of natural language processing is the finding of
grammatical relationships such as subject, object, modifier, etc. Such
relationships are the objects of study in relational grammar
\cite{Perlmutter83}. Many systems (e.g., the KERNEL system
\cite{PPWF93}) use these relationships as an intermediate form when
determining the semantics of syntactically parsed text. In the SPARKLE
project \cite{Sparkle1}, grammatical relations form the layer above
the phrasal-level in a three layer syntax scheme.  Grammatical
relationships are often stored in some type of structure like the
F-structures of lexical-functional grammar \cite{Kaplan94}.

Our own interest in grammatical relations is as a semantic basis for
information extraction in the {\em Alembic} system. The extraction
approach we are currently investigating exploits grammatical relations
as an intermediary between surface syntactic phrases and propositional
semantic interpretations. By directly associating syntactic heads with
their arguments and modifiers, we are hoping that these grammatical
relations will provide a high degree of generality and reliability to
the process of composing semantic representations. This ability to
``parse'' into a semantic representation is according to Charniak
\cite[p.~42]{Charniak97}, ``the most important task to be tackled
now.''

In this paper, we describe a system to learn rules for finding
grammatical relationships when just given a partial parse with
entities like names, core noun and verb phrases (noun and verb groups)
and semi-accurate estimates of the attachments of prepositions and
subordinate conjunctions.  In our system, the different entities,
attachments and relationships are found using rule sequence processors
that are cascaded together.  Each processor can be thought of as
approximating some aspect of the underlying grammar by finite-state
transduction.

We present the problem scope of interest to us, as well as the data
annotations required to support our investigation.  We also present a
decision procedure for finding grammatical relationships.  In brief,
on our training and test set, our procedure achieves 63.6\% recall and
77.3\% precision, for an f-score of 69.8.

\section{Phrase Structure and Grammatical Relations}

In standard derivational approaches to syntax, starting as early as
1965 \cite{Chomsky65}, the notion of grammatical relationship is
typically parasitic on that of phrase structure.  That is to say, the
primary vehicles of syntactic analysis are phrase structure trees;
grammatical relationships, if they are to be considered at all, are
given as a secondary analysis defined in terms of phrase structure.
The surface subject of a sentence, for example, is thus no more than
the NP attached by the production \mbox{S $\rightarrow$ NP VP}; i.e.,
it is the left-most NP daughter of an S node.

The present paper takes an alternate outlook.  In our current work,
grammatical relationships play a central role, to the extent even of
replacing phrase structure as the descriptive vehicle for many
syntactic phenomena.  To be specific, our approach to syntax operates
at two levels: (1) that of core phrases, which are analyzed through
standard derivational syntax, and (2) that of argument and modifier
attachments, which are analyzed through grammatical relations. These
two levels roughly correspond to the top and bottom layers of the
three layer syntax annotation scheme in the SPARKLE project
\cite{Sparkle1}.

\subsection{Core syntactic phrases}\label{s:groups}

In recent years, a consensus of sorts has emerged that postulates some
core level of phrase analysis.  By this we mean the kind of
non-recursive simplifications of the NP and VP that in the literature
go by names such as noun/verb groups \cite{Appelt93}, chunks
\cite{Abney96}, or base NPs \cite{RandM95}.

The common thread between these approaches and ours is to approximate
full noun phrases or verb phrases by only parsing their non-recursive
core, and thus not attaching modifiers or arguments.  For English noun
phrases, this amounts to roughly the span between the determiner and
the head noun; for English verb phrases, the span runs roughly from
the auxiliary to the head verb.  We call such simplified syntactic
categories {\em groups}, and consider in particular, noun, verb,
adverb, adjective, and {\em IN} groups.\footnote{In addition, for the
noun group, our definition encompasses the named entity task, familiar
from information extraction \cite{Muc6}.  Named entities include among
others the names of people, places, and organizations, as well as
dates, expressions of money, and (in an idiosyncratic extension)
titles, job descriptions, and honorifics.} An {\em IN}
group\footnote{The name comes from the Penn Treebank part-of-speech
label for prepositions and subordinate conjunctions.} contains a
preposition or subordinate conjunction (including {\em wh}-words and
``that'').

For example, for \mbox{\em ``I saw the cat that ran.''}, we have the
following core phrase analysis:
\begin{center}
[I]$_{ng}$ [saw]$_{vg}$ [the cat]$_{ng}$ [that]$_{ig}$ [ran]$_{vg}$.
\end{center}
where [...]$_{ng}$ indicates a noun group, [...]$_{vg}$ a verb group,
and [...]$_{ig}$ an IN group.

In English and other languages where core phrases (groups) can be
analyzed by head-out (island-like) parsing, the group head-words
are basically a by-product of the core phrase analysis.

Distinguishing core syntax groups from traditional syntactic phrases
(such as NPs) is of interest because it singles out what is usually
thought of as easy to parse, and allows that piece of the parsing
problem to be addressed by such comparatively simple means as
finite-state machines or transformation sequences.  What is then left
of the parsing problem is the difficult stuff: namely the attachment
of prepositional phrases, relative clauses, and other constructs that
serve in modification, adjunctive, or argument-passing roles.

\subsection{Grammatical relations}

In the present work, we encode this hard stuff through a small
repertoire of {\em grammatical relations}.  These relations hold
directly between constituents, and as such define a graph, with core
constituents as nodes in the graph, and relations as labeled arcs.
Our previous example, for instance, generates the following
grammatical relations graph (head words underlined):
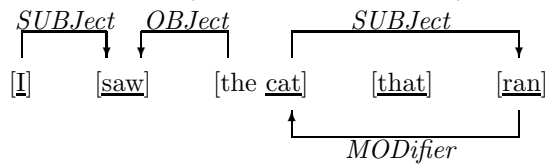
\begin{center}
\begin{picture}(200,50)(-100,-25)
\put(-97,20){\line(0,-1){10}}\put(-97,20){\line(1,0){32}}\put(-81,21){\makebox(0,0)[b]{\em SUBJect}}\put(-65,20){\vector(0,-1){10}}
\put(-52,20){\vector(0,-1){10}}\put(-52,20){\line(1,0){33}}\put(-35,21){\makebox(0,0)[b]{\em OBJect}}\put(-19,20){\line(0,-1){10}}
\put(5,20){\line(0,-1){10}}\put(5,20){\line(1,0){86}}\put(46,21){\makebox(0,0)[b]{\em SUBJect}}\put(91,20){\vector(0,-1){10}}
\put(0,0){\makebox(0,0){[\underline{I}]\hspace{2em} [\underline{saw}]\hspace{2em}
[the \underline{cat}]\hspace{2em} [\underline{that}]\hspace{2em} [\underline{ran}]}}
\put(5,-20){\vector(0,1){10}}\put(5,-20){\line(1,0){86}}\put(46,-21){\makebox(0,0)[t]{\em MODifier}}\put(91,-20){\line(0,1){10}}
\end{picture}
\end{center}

Our grammatical relations effectively replace the recursive $\overline{X}$
analysis of traditional phrase structure grammar.  In this respect,
the approach bears resemblance to a dependency grammar, in that it
has no notion of a spanning S node, or of intermediate constituents
corresponding to argument and modifier attachments.

One major point of departure from dependency grammar, however, is that
these grammatical relation graphs can generally not be reduced to
labeled trees.  This happens as a result of argument passing, as in
\begin{center}
{\em [Fred] [promised] [to help] [John]}
\end{center}
where {\em [Fred]} is both the subject of {\em [promised]} and {\em
[to help]}. This also happens as a result of argument-modifier cycles, as in
\begin{center}
{\em [I] [saw] [the cat] [that] [ran]}
\end{center}
where the relationships between \mbox{\em [the cat]} and {\em [ran]}
form a cycle: \mbox{\em [the cat]} has a subject
relationship/dependency to {\em [ran]}, and {\em [ran]} has a modifier
dependency to \mbox{\em [the cat]}, since {\em [ran]} helps indicate
(modifies) which cat is seen.

There has been some work at making additions to extract grammatical
relationships from a dependency tree structure \cite{Broker98,LandH98}
so that one first produces a surface structure dependency tree with a
syntactic parse and then extracts grammatical relationships from that
tree. In contrast, we skip trying to find a surface structure tree and
just proceed to more directly finding the grammatical relationships,
which are the relationships of interest to us.

A reason for skipping the tree stage is that extracting grammatical
relations from a surface structure tree is often a nontrivial task by
itself. For instance, the precise relationship holding between two
constituents in a surface structure tree cannot be derived
unambiguously from their relative attachments.  Contrast, for example
{\em ``the attack on the military base''} with {\em ``the attack on
March 24''}.  Both of these have the same underlying surface structure
(a PP attached to an NP), but the former encodes the direct object of
a verb nominalization, while the latter encodes a time modifier.
Also, in a surface structure tree, long-distance dependencies between
heads and arguments are not explicitly indicated by attachments
between the appropriate parts of the text.  For instance in {\em
``Fred promised to help John''}, no direct attachment exists between
the {\em ``Fred''} in the text and the {\em ``help''} in the text,
despite the fact that the former is the subject of the latter.

For our purposes, we have delineated approximately a dozen
head-to-argument relationships as well as a commensurate number of
modification relationships. 
Among the head-to-argument relationships, we have the deep subject and
object (SUBJ and OBJ respectively), and also include the surface
subject and object of copulas (COP-SUBJ and the various COP-OBJ
forms).  In addition, we include a number of relationships (e.g.,
PP-SUBJ, PP-OBJ) for arguments that are mediated by prepositional
phrases. An example is in
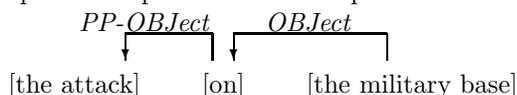
\begin{center}
\begin{picture}(200,30)(-100,-5)
\put(-52,20){\vector(0,-1){10}}\put(-52,20){\line(1,0){33}}\put(-45,21){\makebox(0,0)[b]{\em PP-OBJect}}\put(-19,20){\line(0,-1){10}}
\put(-11,20){\vector(0,-1){10}}\put(-11,20){\line(1,0){58}}\put(17,21){\makebox(0,0)[b]{\em OBJect}}\put(47,20){\line(0,-1){10}}
\put(0,0){\makebox(0,0){[the attack]\hspace{2em} [on]\hspace{2em} [the military base]}}
\end{picture}
\end{center}
where \mbox{\em [the attack]}, a noun group with a verb
nominalization, has its object \mbox{\em [the military base]} passed
to it via the preposition in {\em [on]}.  Among modifier
relationships, we designate both generic modification and some
specializations like locational and temporal modification.  A complete
definition of all the grammatical relations is beyond the scope of
this paper, but we give a summary of usage in Table~\ref{t:g-r-l}.  An
earlier version of the definitions can be found in our annotation
guidelines \cite{Ferro98}. The appendix shows some examples of
grammatical relationship labeling from our experiments.
\begin{table*}[htbp]
\begin{center}
\begin{tabular}{|l|l|l|} \hline
\multicolumn{2}{|c|}{RELATION} & EXAMPLE(s) in the format\\
Name & Description & {\em [source] $\rightarrow$ [target]} in ``text''\\\hline
subj & subject & {\em [I] $\rightarrow$ [promised]} in ``I promised to help''\\
&--- subject of a verb & {\em [I] $\rightarrow$ [to help]} in ``I promised to help''\\
&& {\em [the cat] $\rightarrow$ [ran]} in ``the cat that ran''\\
&--- link a copula subject and object& {\em [You] $\rightarrow$ [happy]} in ``You are happy''\\
&& {\em [You] $\rightarrow$ [a runner]} in ``You are a runner''\\
&--- link a state with the item in that state& {\em [you] $\rightarrow$ [happy]} in ``They made you happy''\\
&--- link a place with the item moving & {\em [I] $\rightarrow$ [home]} in ``I went home''\\
&\hspace{2.5em}to or from that place&\\\hline
obj & object & {\em [saw] $\leftarrow$ [the cat]} in ``I saw the cat''\\
&--- object of a verb & {\em [promised] $\leftarrow$ [to help]} in ``I promised to help you''\\
&--- object of an adjective & {\em [happy] $\leftarrow$ [to help]} in ``I was happy to help''\\
&--- surface subject in passives & {\em [I] $\rightarrow$ [was seen]} in ``I was seen by a cat''\\
&--- object of a preposition, & {\em [by] $\leftarrow$ [the tree]} in ``I was by the tree''\\
&\hspace{2.5em}not for partitives or subsets&\\
&--- object of & {\em [After] $\leftarrow$ [left]} in ``After I left, I ate''\\
&\hspace{2.5em}an adverbial clause complementizer&\\\hline
loc-obj & location object & {\em [went] $\leftarrow$ [home]} in ``I went home''\\
& --link a movement verb with a place & {\em [went] $\leftarrow$ [in]} in ``I went in the house\\
&\hspace{1.5em}where entities are moving to or from &\\\hline
indobj &indirect object& {\em [gave] $\leftarrow$ [you]} in ``I gave you a cake''\\\hline
empty & use instead of ``subj'' relation when subject & {\em [There] $\rightarrow$ [trees]} in ``There are trees'' \\
& is an expletive (existential) ``it'' or ``there''&  \\\hline
pp-subj & genitive functional ``of'''s & {\em [name] $\leftarrow$ [of]} in ``name of the building''\\
&use instead of ``subj'' relation when the& {\em [was seen] $\leftarrow$ [by]} in ``I was seen by a cat''\\
&\hspace{1em}subject is linked via a preposition, &\\
&\hspace{1em}links preposition to its head &\\\hline
pp-obj & nongenitive functional ``of'''s & {\em [age] $\leftarrow$ [of]} in ``age of 12''\\
& use in place of ``obj'' relation when the & {\em [the attack] $\leftarrow$ [on]} in ``the attack on the base''\\
&\hspace{1em}object is linked via a preposition, &\\
&\hspace{1em}links preposition to its head &\\\hline
pp-io &use in place of ``indobj'' relation when the & {\em [gave] $\leftarrow$ [to]} in ``gave a cake to them''\\
&\hspace{1em}indirect object is linked via a preposition, &\\
&\hspace{1em}links preposition to its head &\\\hline
cop-subj &surface subject for a copula& {\em [You] $\rightarrow$ [are]} in ``You are happy''\\\hline
n-cop-obj &surface nominative object for a copula&{\em [is] $\leftarrow$ [a rock]} in ``It is a rock''\\\hline
p-cop-obj &surface predicate object for a copula & {\em [are] $\leftarrow$ [happy]} in ``You are happy''\\\hline
subset & subset&{\em [five] $\rightarrow$ [the kids]} in ``five of the kids''\\\hline
mod &generic modifier (use when &{\em [the cat] $\leftarrow$ [ran]} in ``the cat that ran''\\
&\hspace{1.5em}modifier does not fit in a case below)&{\em [ran] $\leftarrow$ [with]} in ``I ran with new shoes''\\\hline
mod-loc &location modifier& {\em [ate] $\leftarrow$ [at]} in ``I ate at home''  \\\hline
mod-time &time modifier&{\em [ate] $\leftarrow$ [at]} in ``I ate at midnight''\\
&&{\em [Yesterday] $\rightarrow$ [ate]} in ``Yesterday, I ate''\\\hline
mod-poss &possessive modifier&{\em [the cat] $\rightarrow$ [toy]} in ``the cat's toy'' \\\hline
mod-quant &quantity modifier (partitive) & {\em [hundreds] $\rightarrow$ [people]} in ``hundreds of people''\\\hline
mod-ident &identity modifier (names)& {\em [a cat] $\leftarrow$ [Fuzzy]} in ``a cat named Fuzzy''\\
&&{\em [the winner] $\leftarrow$ [Pat Kay]} \\
&&\hspace{2.5em}in ``the winner, Pat Kay, is''\\\hline
mod-scalar &scalar modifier&{\em [16 years] $\rightarrow$ [ago]} in ``16 years ago''\\\hline
\end{tabular}
\end{center}
\caption{Summary of grammatical relationships}\label{t:g-r-l}
\end{table*}

Our set of relationships is similar to the set used in the SPARKLE
project \cite{Sparkle1} \cite{CBS98}. One difference is that we make
many semantically-based distinctions between what SPARKLE calls a {\bf
mod}ifier, such as time and location modifiers, and the various
arguments of event nouns.

\subsection{Semantic interpretation}

A major motivation for this approach is that it supports a direct
mapping into semantic interpretations. In our framework, semantic
interpretations are given in a neo-Davidsonian propositional
logic. Grammatical relations are thus interpreted in terms of mappings
and relationships between the constants and variables of the
propositional language. For instance, the deep subject relation (SUBJ)
maps to the first position of a predicate's argument list, the deep
object (OBJ) to the second such position, and so forth.

Our example sentence, {\em ``I saw the cat that ran''} thus translates
directly to the following:
\begin{center}
\begin{tabular}{|l|l|} \hline
Proposition & Comment\\\hline
saw(x1 x2)  &  SUBJ and OBJ relations\\
I(x1)       &\\
cat(x2)     &\\         
ran(x2)=e3  & SUBJ relation\\
& \hspace{1em}(e3 is the event variable)\\
mod(e3 x2)  & MOD relation\\\hline
\end{tabular}
\end{center}

We do not have an explicit level for clauses between our core phrase
and grammatical relations levels. However, we do have a set of
implicit clauses in that each verb (event) and its arguments can be
deemed a base level clause. In our example {\em ``I saw the cat that
ran''}, we have two such base level clauses. {\em ``saw''} and its
arguments form the clause {\em ``I saw the cat''}. {\em ``ran''} and
its argument form the clause {\em ``the cat ran''}.  Each noun with a
possible semantic class of ``act'' or ``process'' in Wordnet
\cite{Wordnet} (and that noun's arguments) can likewise be deemed a
base level clause.

\section{The Processing Model}\label{s:processing-model}
Our system uses transformation-based error-driven learning to
automatically learn rules from training examples \cite{BandR94}.

One first runs the system on a training set, which starts with no
grammatical relations marked. This training run moves in iterations,
with each iteration producing the next rule that yields the best net
gain in the training set (number of matching relationships found minus
the number of spurious relationships introduced). On ties, rules with
less conditions are favored over rules with more conditions. The
training run ends when the next rule found produces a net gain below a
given threshold.

The rules are then run in the same order on the test set to see how
well they do.

The rules are condition/action pairs that are tried on each syntax
group. The actions in our system are limited to attaching (or
unattaching) a relationship of a particular type from the group under
consideration to that group's neighbor a certain number of groups away
in a particular direction (left or right). A sample action would be to
attach a SUBJ relation from the group under consideration
to the group two groups away to the right.

A rule only applies to a syntax group when that group and its
neighbors meet the rule's conditions. Each condition tests the group
in a particular position relative to the group under consideration
(e.g., two groups away to the left). All tests can be negated.
Table~\ref{t:conditions} shows the possible tests.
\begin{table*}[htb]
\begin{center}
\begin{tabular}{|l|l|} \hline
Test Type & Example, Sample Value(s) \\\hline 
group type & noun, verb \\\hline
verb group property & passive, infinitival, \\
 & unconjugated present participle \\\hline 
end group in a sentence & first, last \\\hline
pp-attachment & Is a preposition or subordinate \\
 &conjunction attached to the \\
 & group under consideration? \\\hline
group contains & a particular lexeme or part-of-speech \\\hline
between two groups, there is & a particular lexeme or part-of-speech \\\hline
group's head (main) word & ``cat'' \\\hline
head word part-of-speech & common plural noun \\\hline
head word within a named entity & person, organization \\\hline
head word subcategorization and complement categories & intransitive verbs \\ 
(from Comlex \cite{Comlex}, over 100 categories) & \\\hline
head word semantic classes & process, communication \\
(from Wordnet \cite{Wordnet}, 25 noun and 15 verb classes)& \\\hline
punctuation or coordinating conjunction & exist between two groups? \\\hline
head word in a word list? & list of relative pronouns,\\
 & list of partitive quantities (e.g., {\em ``some''})\\\hline
\end{tabular}
\end{center}
\caption{Possible tests}\label{t:conditions}
\end{table*}

A sample rule is when a noun group $n$'s
\begin{itemize}
\item immediate group to the right has some form of the verb {\em
``be''} as the head-word,
\item immediate group to the left is not an IN group (preposition, {\em
wh}-word, etc.) and
\item $n$'s head-word is not an existential {\em ``there''}
\end{itemize}
make $n$ a SUBJ of the group two groups over to $n$'s right.

When applied to the group \mbox{\em [The \underline{cat}]} (head words
are underlined) in the sentence
\begin{center}
{\em [The \underline{cat}] [\underline{was}] [very \underline{happy}].}
\end{center}
this rule makes \mbox{\em [The \underline{cat}]} a SUBJect of
\mbox{\em [very \underline{happy}]}.

Searching over the space of possible rules is very computationally
expensive. Our system has features to make it easier to perform
searching in parallel and to minimize the amount of work that needs to
be undone once a rule is selected. With these features, rules that
(un)attach different types of relationships or relationships at
different distances can be searched independently of each other in
parallel.

One feature is that the action of any rule only affects the
applicability of rules with either the exact same or opposite
action. For example, selecting and running a rule which attaches a MOD
relationship to the group that is two groups to the right only can
affect the applicability of other rules that either attach or unattach
a MOD relationship to the group that is two groups to the right.

Another feature is the use of net gain as a proxy measure during
training.  The actual measure by which we judge the system's
performance is called an {\em f-score}. This {\em f-score} is a type
of harmonic mean of the precision ($p$) and recall ($r$) and is given
by \mbox{$2pr/(p+r)$}. Unfortunately, this measure is nonlinear, and
the application of a new rule can alter the effects of all other
possible rules on the {\em f-score}. To enable the described parallel
search to take place, we need a measure in which how a rule affects
that measure only depends on other rules with either the exact same or
opposite action. The net gain measure has this trait, so we use it as
a proxy for the {\em f-score} during training.


Another way to increase the learning speed is to restrict the number
of possible combinations of conditions/constraints or actions to
search over. Each rule is automatically limited to only considering
one type of syntactic group. Then when searching over possible
conditions to add to that rule, the system only needs to consider the
parts-of-speech, semantic classes, etc. applicable to that type of
group.

Many other restrictions are possible. One can estimate which
restrictions to try by making some training and test runs with
preliminary data sets and seeing what restrictions seem to have no
effect on performance, etc. The restrictions used in our experiments
are described below.

\section{Experiments}
\subsection{The Data}
Our data consists of bodies of some elementary school reading
comprehension tests. For our purposes, these tests have the advantage
of having a fairly predictable size (each body has about 100
relationships and syntax groups) and a consistent style of
writing. The tests are also on a wide range of topics, so we avoid a
narrow specialized vocabulary. Our training set has 1963 relationships
(2153 syntax groups, 3299 words) and our test set has 748
relationships (830 syntax groups, 1151 words).

We prepared the data by first manually removing the headers and the
questions at the end for each test. We then manually annotated the
remainder for named entities, syntax groups and relationships.  As the
system reads in our data, it automatically breaks the data into
lexemes and sentences, tags the lexemes for part-of-speech and
estimates the attachments of prepositions and subordinate
conjunctions. The part-of-speech tagging uses a high-performance
tagger based on \cite{BrillPhD}. The attachment estimation uses a
procedure described in \cite{YandV98} when multiple left attachment
possibilities exist and four simple rules when no or only one left
attachment possibility exists. Previous testing indicates that the
estimation procedure is about 75\% accurate.

\subsection{Parameter Settings for Training}\label{s:train-parm-set}
As described earlier, a training run uses many parameter
settings. Examples include where to look for relationships and to test
conditions, the maximum number of constraints allowed in a rule, etc.

Based on the observation that 95\% of the relationships are to at most
three groups away in the training set, we decided to limit the search for
relationships to at most three groups in length. To keep the number of
possible constraints down, we disallowed the negations of most tests
for the presence of a particular lexeme or lexeme stem.

To help determine many of the settings, we made some preliminary runs
using different subsets of our final training set as the preliminary
training and test sets. This kept the final test set unexamined during
development. From these preliminary runs, we decided to limit a rule
to at most three constraints\footnote{In addition to the constraint on
the relationship's source group type.} in order to keep the training
time reasonable.  We found a number of limitations that help speed up
training and seemed to have no effect on the preliminary test runs. A
threshold of four was set to end a training run. So training ends when
it can no longer find a rule that produces at least a net gain of four
in the score.  Only syntax groups spanned by the relationship being
attached or unattached and those groups' immediate neighbors were
allowed to be mentioned in a rule's conditions. Each condition testing
a head-word had to test a head-word of a different group. Except for
the lexemes ``of'', ``?'' and a few determiners like {\em ``the''},
tests for single lexemes were removed. Also disallowed were negations
of tests for the presence of a particular part-of-speech anywhere
within a syntax group.

In our preliminary runs, lowering the threshold tended to raise recall
and lower precision.

\subsection{The Results}
Training produced a sequence of 95 rules which had 63.6\% recall and
77.3\% precision for an f-score of 69.8 when run on the test set. In
our test set, the key relationships, SUBJ and OBJ, formed the bulk of
the relationships (61\%). Both recall and precision for both SUBJ and
OBJ were above 70\%, which pleased us. Because of their relative
abundance in the test set, these two relationships also had the most
number of errors in absolute terms. Combined, the two accounted for
45\% of the recall errors and 66\% of the precision errors. In terms
of percentages, recall was low for many of the less common
relationships, such as generic, time and location {\em modification}
relationships. In addition, the relative precision was low for those
modification relationships. The appendix shows some examples of our
system responding to the test set.

To see how well the rules, which were trained on reading comprehension
test bodies, would carry over to other texts of non-specialized
domains, we examined a set of six broadcast news stories. This set had
525 relationships (585 syntax groups, 1129 words). By some measures,
this set was fairly similar to our training and test sets. In all three
sets, 33--34\% of the relationships were OBJ and 26--28\% were
SUBJ. The broadcast news set did tend to have relationships
between groups that were slightly further apart:\\
\begin{tabular}{|l|r|r|r|} \hline
& \multicolumn{3}{l|}{Percent of Relations with Length}\\
Set & $\leq 1$ & $\leq 2$ & $\leq 3$ \\\hline
training & 66\% & 87\% & 95\% \\\hline
test & 68\% & 89\% & 96\% \\\hline
broadcast news & 65\% & 84\% & 90\% \\\hline
\end{tabular}\\
This tendency, plus differences in the relative proportions of various
modification relationships are probably what produced the drop in
results when we tested the rules against this news set: recall at
54.6\%, precision at 70.5\% (f-score at 61.6\%).

To estimate how fast the results would improve by adding more training
data, we had the system learn rules on a new smaller training set and
then tested against the regular test set. Recall dropped to 57.8\%,
precision to 76.2\%. The smaller training set had 981 relationships
(50\% of the original training set). So doubling the training data
here (going from the smaller to the regular training set) reduced the
smaller training set's recall error of 42.2\% by 14\% and the
precision error of 23.8\% by 5\%. Using the broadcast news set as a
test produced similar error reduction results.

One complication of our current scoring scheme is that identifying a
{\em modification} relationship and mis-typing it is more harshly
penalized than not finding a {\em modification} relationship at
all. For example, finding a {\em modification} relationship, but
mistakingly calling it a generic modifier instead of a time modifier
produces both a missed key error (not finding a time modifier) and a
spurious response error (responding with a generic modifier where none
exists). Not finding that {\em modification} relationship at all just
produces a missed key error (not finding a time modifier).  This
complication, coupled with the fact that generic, time and location
modifiers often have a similar surface appearance (all are often
headed by a preposition or a complementizer) may have been responsible
for the low recall and precision scores for these types of
modifiers. Even the training scores for these types of modifiers were
particularly low.
To test how well our system finds these three types of {\em
modification} when one does not care about specifying the sub-type, we
reran the original training and test with the three sub-types merged
into one sub-type in the annotation. With the merging, recall of these
{\em modification} relationships jumped from 27.8\% to
48.9\%. Precision rose from 52.1\% to 67.7\%. Since these {\em
modification} relationships are only about 20\% of all the
relationships, the overall improvement is more modest. Recall rises to
67.7\%, precision to 78.6\% (f-score to 72.6).

Taking this one step further, the LOC-OBJ and various PP-$x$ arguments
also all have both a low recall (below 35\%) in the test and a similar
surface structure to that of generic, time and location
modifiers. When these argument types were merged with the three
modifier types into one combined type, their combined recall was
60.4\% and precision was 81.1\%. The corresponding overall test recall
and precision were 70.7\% and 80.5\%, respectively.

\section{Comparison with Other Work}
At one level, computing grammatical relationships can be seen as a
parsing task, and the question naturally arises as to how well this
approach compares to current state-of-the-art parsers. Direct
performance comparisons, however, are elusive, since parsers are
evaluated on an incommensurate tree bracketing task. For example, the
SPARKLE project \cite{Sparkle1} puts tree bracketing and grammatical
relations in two different layers of syntax. Even if we disregard the
questionable aspects of comparing tree bracketing apples to
grammatical relation oranges, an additional complication is the fact
that our approach divides the parsing task into an easy piece (core
phrase boundaries) and a hard one (grammatical relations). The results
we have presented here are given solely for this harder part, which
may explain why at roughly 70 points of f-score, they are lower than
those reported for current state-of-the-art parsers (e.g., Collins
\cite{Collins97}).

More comparable to our approach are some other grammatical relation
finders. Some examples for English include the English parser used in
the SPARKLE project \cite{Sparkle3-1} \cite{Sparkle3-2} \cite{CMB98} and
the finder built with a memory-based approach \cite{ADK98}. These
relation finders make use of large annotated training data sets and/or
manually generated grammars and rules. Both techniques take much
effort and time. At first glance both of these finders perform better
than our approach. Except for the {\em object} precision score of 77\%
in \cite{ADK98}, both finders have grammatical relation recall and
precision scores in the 80s. But a closer examination reveals that
these results are not quite comparable with ours.
\begin{enumerate}
\item Each system is recovering a different variation of grammatical
relations. As mentioned earlier,
one difference between us and the SPARKLE project is that the latter
ignores many of distinctions that we make for different types of
modifiers. The system in \cite{ADK98} only finds a subset of the
surface subjects and objects.

\item In addition, the evaluations of these two finders produced more
complications. In an illustration of the time consuming nature of
annotating or reannotating a large corpus, the SPARKLE project
originally did not have time to annotate the English test data for
modifier relationships. As a result, the SPARKLE English parser was
originally not evaluated on how well it found modifier relationships
\cite{Sparkle3-2} \cite{CMB98}. The reported results as of 1998 only
apply to the argument (subject, object, etc.) relationships.  Later
on, a test corpus with modifier relationship annotation was
produced. Testing the parser against this corpus produced generally
lower results, with an overall recall, precision and f-score of 75\%
\cite{CMB99}. This is still better than our f-score of 70\%, but not
by nearly as much. This comparison ignores the fact that the results
are for different versions of grammatical relationships and for
different test corpora.

The figures given above were the original (1998) results for the
system in \cite{ADK98}, which came from training and testing on data
derived from the Penn Treebank corpus \cite{PennTreebank} in which the
added null elements (like null subjects) were left in. These null
elements, which were given a {\tt -NONE-} part-of-speech, do not
appear in raw text. Later (1999 results), the system was re-evaluated
on the data with the added null elements removed. The {\em subject}
results declined a little. The {\em object} results declined more,
with the precision now lower than ours (73.6\% versus 80.3\%) and the
f-score not much higher (80.6\% versus 77.8\%). This comparison is
also between results with different test corpora and slightly
different notions of what an {\em object} is.
\end{enumerate}

\section{Summary, Discussion, and Speculation}
In this paper, we have presented a system for finding grammatical
relationships that operates on easy-to-find constructs like noun
groups. The approach is guided by a variety of knowledge sources, such
as readily available lexica\footnote{Resources to find a word's
possible stem(s), semantic class(es) and subcategorization
category(ies).}, and relies to some degree on well-understood
computational infrastructure: a p-o-s tagger and an attachment
procedure for preposition and subordinate conjunctions. In sample
text, our system achieves 63.6\% recall and 77.3\% precision (f-score
= 69.8) on our repertory of grammatical relationships.

This work is admittedly still in relatively early stages. Our training
and test corpora, for instance, are less-than-gargantuan compared to
such collections as the Penn Treebank \cite{PennTreebank}. However,
the fact that we have obtained an f-score of 70 from such sparse
training materials is encouraging. The recent implementation of rapid
annotation tools should speed up further annotation of our own native
corpus.

Another task that awaits us is a careful measurement of interannotator
agreement on our version the grammatical relationships.

We are also keenly interested in applying a wider range of learning
procedures to the task of identifying these grammatical
relations. Indeed, a fine-grained analysis of our development test
data has identified some recurring errors related to the rule sequence
approach. A hypothesis for further experimentation is that these
errors might productively be addressed by revisiting the way we
exploit and learn rule sequences, or by some hybrid approach blending
rules and statistical computations. In addition, since generic, time
and location modifiers, and LOC-OBJ and various PP-$x$ arguments often
have a similar surface appearance, one might first just try to locate
all such entities and then in a later phase try to classify them by
type.

Different applications will need to deal with different styles of text
(e.g., journalistic text versus narratives) and different standards of
grammatical relationships. An additional item of experimentation is to
use our system to adapt other systems, including earlier versions of
our system, to these differing styles and standards.

Like other Brill transformation rule systems \cite{BandR94}, our
system can take in the output of another system and try to improve on
it. This suggests a relatively low expense method to adapt a
hard-to-alter system that performs well on a slightly different style
or standard. Our training approach accepts as a starting point an
initial labeling of the data. So far, we have used an empty
labeling. However, our system could just as easily start from a
labeling produced as the output of the hard-to-alter system. The
learning would then not be reducing the error between an empty
labeling and the key annotations, but between the hard-to-alter
system's output and the key annotations. By using our system in this
post-processing manner, we could use a relatively small retraining set to
adapt, for example, the SPARKLE English parser, to our standard of
grammatical relationships without having reengineer that parser.
Palmer \cite{Palmer97} used a similar approach to improve on existing
word segmenters for Chinese. Trying this suggestion out is also
something for us to do.

This discussion of training set size brings up perhaps the most
obvious possible improvement. Namely, enlarging our very small
training set. As has been mentioned, we have recently improved our
annotation environment and look forward to working with more data.

Clearly we have many experiments ahead of us. But we believe that the
results obtained so far are a promising start, and the potential
rewards of the approach are very significant indeed.

\appendix
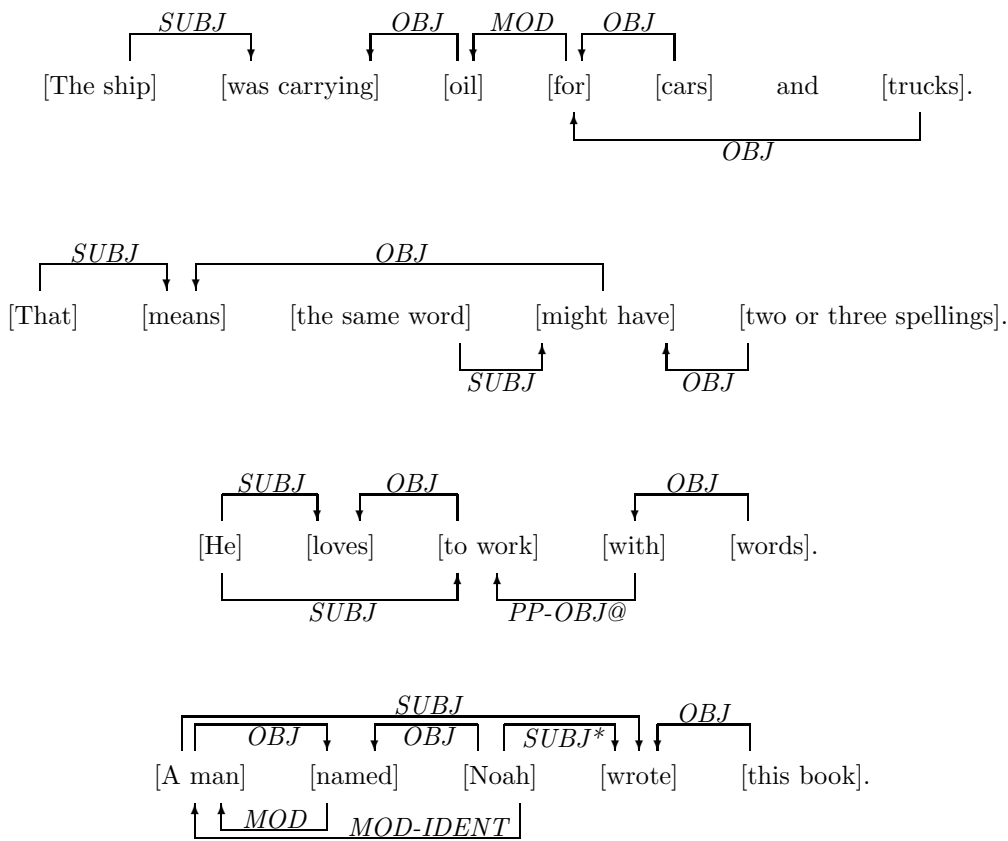
\begin{figure*}[htb]
\begin{center}
\begin{picture}(200,50)(-100,-25)
\put(-143,20){\line(0,-1){10}}\put(-143,20){\line(1,0){46}}\put(-120,21){\makebox(0,0)[b]{\em SUBJ}}\put(-97,20){\vector(0,-1){10}}
\put(-52,20){\vector(0,-1){10}}\put(-52,20){\line(1,0){33}}\put(-35,21){\makebox(0,0)[b]{\em OBJ}}\put(-19,20){\line(0,-1){10}}
\put(-13,20){\vector(0,-1){10}}\put(-13,20){\line(1,0){35}}\put(5,21){\makebox(0,0)[b]{\em MOD}}\put(22,20){\line(0,-1){10}}
\put(28,20){\vector(0,-1){10}}\put(28,20){\line(1,0){35}}\put(45,21){\makebox(0,0)[b]{\em OBJ}}\put(63,20){\line(0,-1){10}}
\put(0,0){\makebox(0,0){[The ship]\hspace{2em} [was carrying]\hspace{2em} [oil]\hspace{2em} [for]\hspace{2em} [cars]\hspace{2em} and\hspace{2em} [trucks].}}
\put(25,-20){\vector(0,1){10}}\put(25,-20){\line(1,0){131}}\put(90,-21){\makebox(0,0)[t]{\em OBJ}}\put(156,-20){\line(0,1){10}}
\end{picture}
\vspace{0.5in}

\begin{picture}(200,50)(-100,-25)
\put(-177,20){\line(0,-1){10}}\put(-177,20){\line(1,0){48}}\put(-153,21){\makebox(0,0)[b]{\em SUBJ}}\put(-129,20){\vector(0,-1){10}}
\put(-118,20){\vector(0,-1){10}}\put(-118,20){\line(1,0){154}}\put(-41,21){\makebox(0,0)[b]{\em OBJ}}\put(36,20){\line(0,-1){10}}
\put(0,0){\makebox(0,0){[That]\hspace{2em} [means]\hspace{2em} [the same word]\hspace{2em} [might have]\hspace{2em} [two or three spellings].}}
\put(-18,-20){\line(0,1){10}}\put(-18,-20){\line(1,0){31}}\put(-3,-21){\makebox(0,0)[t]{\em SUBJ}}\put(13,-20){\vector(0,1){10}}
\put(60,-20){\vector(0,1){10}}\put(60,-20){\line(1,0){31}}\put(75,-21){\makebox(0,0)[t]{\em OBJ}}\put(91,-20){\line(0,1){10}}
\end{picture}
\vspace{0.5in}

\begin{picture}(200,50)(-100,-25)
\put(-108,20){\line(0,-1){10}}\put(-108,20){\line(1,0){36}}\put(-90,21){\makebox(0,0)[b]{\em SUBJ}}\put(-72,20){\vector(0,-1){10}}
\put(-56,20){\vector(0,-1){10}}\put(-56,20){\line(1,0){37}}\put(-37,21){\makebox(0,0)[b]{\em OBJ}}\put(-19,20){\line(0,-1){10}}
\put(48,20){\vector(0,-1){10}}\put(48,20){\line(1,0){43}}\put(69,21){\makebox(0,0)[b]{\em OBJ}}\put(91,20){\line(0,-1){10}}
\put(0,0){\makebox(0,0){[He]\hspace{2em} [loves]\hspace{2em} [to work]\hspace{2em} [with]\hspace{2em} [words].}}
\put(-108,-20){\line(0,1){10}}\put(-108,-20){\line(1,0){89}}\put(-63,-21){\makebox(0,0)[t]{\em SUBJ}}\put(-19,-20){\vector(0,1){10}}
\put(-4,-20){\vector(0,1){10}}\put(-4,-20){\line(1,0){52}}\put(22,-21){\makebox(0,0)[t]{\em PP-OBJ@}}\put(48,-20){\line(0,1){10}}
\end{picture}
\vspace{0.5in}

\begin{picture}(200,50)(-100,-25)
\put(-125,23){\line(0,-1){13}}\put(-125,23){\line(1,0){173}}\put(-32,24){\makebox(0,0)[b]{\em SUBJ}}\put(48,23){\vector(0,-1){13}}
\put(-120,20){\line(0,-1){10}}\put(-120,20){\line(1,0){50}}\put(-91,19){\makebox(0,0)[t]{\em OBJ}}\put(-70,20){\vector(0,-1){10}}
\put(-52,20){\vector(0,-1){10}}\put(-52,20){\line(1,0){39}}\put(-32,19){\makebox(0,0)[t]{\em OBJ}}\put(-13,20){\line(0,-1){10}}
\put(-3,20){\line(0,-1){10}}\put(-3,20){\line(1,0){42}}\put(19,19){\makebox(0,0)[t]{\em SUBJ*}}\put(39,20){\vector(0,-1){10}}
\put(55,20){\vector(0,-1){10}}\put(55,20){\line(1,0){35}}\put(72,21){\makebox(0,0)[b]{\em OBJ}}\put(90,20){\line(0,-1){10}}
\put(0,0){\makebox(0,0){[A man]\hspace{2em} [named]\hspace{2em}
[Noah]\hspace{2em} [wrote]\hspace{2em} [this book].}}
\put(-110,-20){\vector(0,1){10}}\put(-110,-20){\line(1,0){40}}\put(-90,-19){\makebox(0,0)[b]{\em MOD}}\put(-70,-20){\line(0,1){10}}
\put(-120,-23){\vector(0,1){13}}\put(-120,-23){\line(1,0){123}}\put(-32,-22){\makebox(0,0)[b]{\em MOD-IDENT}}\put(3,-23){\line(0,1){13}}
\end{picture}
\end{center}
\caption{Example test responses from our system. @ marks the missed key. *
marks the spurious response.}\label{f:res-1}
\end{figure*}
\section{Appendix: Examples from Test Results}
Figure~\ref{f:res-1} shows some example sentences from the test
results of our main experiment.\footnote{The material came from level
2 of ``The 5 W's'' written by Linda Miller. It is available from
Remedia Publications, 10135 E. Via Linda \#D124, Scottsdale, AZ 85258,
USA.} @ marks the relationship that our system missed. * marks the
relationship that our system wrongly hypothesized. In these examples,
our system handled a number of phenomena correctly, including:
\begin{itemize}
 \item The coordination conjunction of the objects \\
{\em [cars] and [trucks]}

 \item The verb group \mbox{\em [might have]} being an object of another
 verb. 

 \item The noun group {\em [He]} being the subject of two verbs.

 \item The relationships within the reduced relative clause\\
{\em [A man] [named] [Noah]}, which makes one noun group a name or
label for another noun group. 
\end{itemize}
Our system misses a PP-OBJ relationship, which is a low occurrence
relationship. Our system also accidentally make both {\em [A man]} and
{\em [Noah]} subjects of the group {\em [wrote]} when only the former
should be.

\end{document}